\documentclass{article}

\usepackage{amsmath}
\usepackage{amssymb}
\usepackage{algorithm}
\usepackage{algpseudocode}

\usepackage{epsfig}
\usepackage{graphicx}
\usepackage{subcaption}

\usepackage{booktabs}

\usepackage[dvipsnames]{xcolor}

\newcommand{\comment}[1]{}

\usepackage[final]{corl_2020} 

\title{Learning Obstacle Representations\\ for Neural Motion Planning}

\author{
  Robin Strudel\\
  Inria\begin{NoHyper}\thanks{Inria, \'Ecole normale supérieure, CNRS, PSL Research University, 75005 Paris, France.}\end{NoHyper}\\
  \And
  Ricardo Garcia \\
  Inria\begin{NoHyper}\thanks{University Grenoble Alpes, Inria, CNRS, Grenoble INP, LJK, 38000 Grenoble, France.}\end{NoHyper}\\
  \And
  Justin Carpentier \\
  Inria\footnotemark[1]\\
  \And
  Jean-Paul Laumond \\
  Inria\footnotemark[1]\\
  \And
  Ivan Laptev \\
  Inria\footnotemark[1]
  \And
  Cordelia Schmid \\
  Inria\footnotemark[1]
}

\begin{document}
\maketitle

\vspace{-0.7cm}
\begin{abstract}
Motion planning and obstacle avoidance is a key challenge in robotics applications.
While previous work succeeds to provide excellent solutions for known environments, sensor-based motion planning in new and dynamic environments remains difficult. 
In this work we address sensor-based motion planning from a learning perspective.
Motivated by recent advances in visual recognition, we argue the importance of learning appropriate representations for motion planning.
We propose a new obstacle representation based on the PointNet architecture
and train it jointly with policies for obstacle avoidance. 
We experimentally evaluate our approach for rigid body motion planning in challenging environments and demonstrate significant improvements of the state of the art in terms of accuracy and efficiency. 
\end{abstract}


\section{Introduction}
\vspace{-0.2cm}

Motion planning is a fundamental robotics problem~\citep{Latombe1991,Lavalle2006} with numerous applications in mobile robot navigation \citep{Fox1997}, industrial robotics \citep{Laumond2006}, humanoid robotics \citep{Harada2014} and other domains. Sampling-based methods such as Rapidly Exploring Random Trees~(RRT)~\citep{Kuffner2000} and Probabilistic Roadmaps~(PRM)~\citep{Kavraki1996} have been shown successful for finding a collision-free path in complex environments with many obstacles.
Such methods are able to solve the so-called piano mover problem~\citep{Sharir1986} and typically assume static environments and prior knowledge about the shape and location of obstacles. 
In many practical applications, however, it is often difficult or even impossible to obtain detailed a-priori knowledge about the real state of environments. 
It is therefore desirable to design methods relying on partial observations obtained from sensor measurements and enabling motion planning in unknown and possibly dynamic environments. 
Moreover, given the high complexity devoted to exploration in sampling-based methods, it is also desirable to design more efficient methods that use prior experience to quickly find solutions for motion planning in new environments.

To address the above challenges, several works~\citep{Glasius1995, Yang2000, Pfeiffer2017, Ichter2018, Jurgenson2019, Qureshi2019} adopt neural networks to learn motion planning from previous observations. Such Neural Motion Planning~(NMP) methods either improve the exploration strategies of sampling-based approaches~\citep{Ichter2018} or learn motion policies with imitation learning~\citep{Pfeiffer2017, Qureshi2019} and reinforcement learning~\citep{Jurgenson2019}. In this work we follow the NMP paradigm and propose a new learnable obstacle representation for motion planning. 

Motivated by recent advances in visual recognition~\citep{Qi2017,krizhevsky2012imagenet,He_2016_CVPR}, we argue that the design and learning of obstacle representations plays a key role for the success of learning-based motion planning. 

\begin{figure*}
  \centering
    \includegraphics[width=\textwidth]{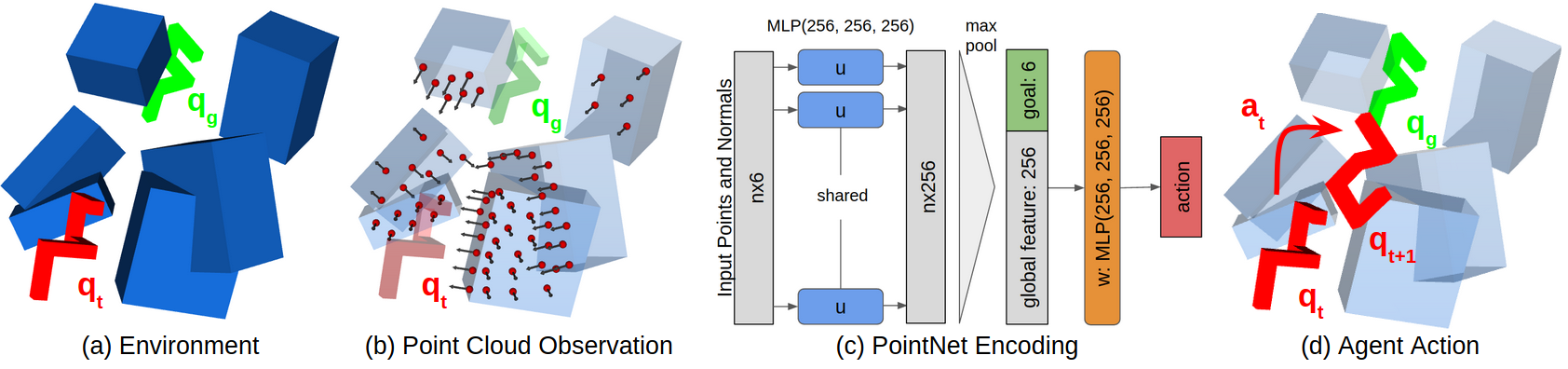}
  \caption{
  Overview of our approach. (a) We aim to find a collision-free path for a rigid body from its current configuration $q_t$ to the goal configuration $q_g$. (b) We assume no prior knowledge about the scene and represent obstacles by points and normals sampled on object surfaces. (c) Our neural network learns the PointNet encoding of observed points and normals together with the motion policy. (d) The learned network generates actions that move the body towards the goal configuration along a collision-free path.   
 }
  \label{fig:overview}
\vspace{-0.4cm}
\end{figure*}

In this work, we proposed a new representation based on point clouds which are first sampled from visible surfaces of obstacles and then encoded with a \mbox{PointNet}~\cite{Qi2017} neural network architecture, see Fig.~\ref{fig:overview}.
We learn our obstacle representation jointly with the motion planing policies in the SAC\,(Soft Actor Critic) reinforcement learning framework~\cite{SAC}.
In particular, while using environments composed of 3D-box shapes during training, we demonstrate the generalization of our motion planning policies to complex environments with objects of previously unseen shapes.
We also show our method to seamlessly generalize to new object constellations and dynamic scenes with moving obstacles.
We evaluate our method in challenging simulated environments and validate our representation through ablation studies and comparison to the state of the art. 
Our method significantly improves the accuracy of previous NMP methods while being one order of magnitude more computationally efficient compared to close competitors.

The contributions of our work are threefold. 
First, we propose a new learnable obstacle representation based on point clouds and the PointNet neural architecture~\citep{Qi2017}.
Second, we learn our representation jointly with policies for rigid body motion planning using reinforcement learning. 
Finally, we experimentally evaluate our approach and demonstrate significant improvements of the state of the art in motion planning in terms of accuracy and efficiency. Code and qualitative results along with a video are available on the project webpage~\citep{nmpwebpage}.

\vspace{-0.4cm}
\section{Related Work}
\label{sec:citations}
\vspace{-0.4cm}

\textbf{Sampling-based motion planning.}
Sampling-based methods such as RRT~\citep{Kuffner2000} have
been extensively studied for motion planning~\citep{Latombe1991, Latombe1998, Lavalle2006, Kuffner2000, Kavraki1996}. 
Such methods can deal with complex environments, however, they typically assume a complete knowledge of the robot workspace and static obstacles. 
Given the complexity of sampling-based methods, recent work proposes efficient exploration schemes~\citep{Jetchev2010, Lien2010, Branicky2008, Martin2007}, e.g., by reusing previously discovered paths in similar scenarios~\citep{Branicky2008} or biasing RRT search to promising regions~\citep{Martin2007}.
Our work does not assume any a-priori knowledge about the environment and can deal with moving obstacles.
Moreover, while our method performs an extensive exploration during training, the learned policies directly generate feasible paths during testing in new environments.

\textbf{Neural motion planning.} Learning-based methods for motion planning have been introduced in~\citep{Glasius1995, Yang2000}.
Motivated by the success of deep learning, a number of more recent methods explore neural networks for motion planning and obstacle avoidance. 
\citet{Ichter2018} learns to sample robot configurations close to the target RRT solutions. 
\citet{Qureshi2019} and \citet{Pfeiffer2017} learn a policy with the imitation
learning using RRT solutions as demonstrations. 
While improving the efficiency of RRT, \citep{Pfeiffer2017,Ichter2018, Qureshi2019}~still require sampling at test time and are not easily applicable in scenes with moving obstacles. 
Our work is most related to~\citet{Jurgenson2019} who explore reinforcement learning (RL) to learn motion policies. Similar to~\citep{Jurgenson2019} we use RL and learn to avoid obstacles using a negative collision reward. While~\citep{Jurgenson2019} presents results in relatively simple 2D environments, we propose a new learnable obstacle representation that generalizes to complex 3D scenes. We experimentally compare our method to~\citep{Ichter2018, Jurgenson2019, Qureshi2019} and demonstrate improved accuracy. 

\textbf{Visual Representation.}
The NMP methods \cite{Jurgenson2019, Qureshi2019, Ichter2018} assume full knowledge of the workspace and use an obstacle encoding either based on a 2D image of obstacles encoded with a convolutional neural network (CNN) \cite{Jurgenson2019}, or an occupancy grid \cite{Ichter2018} or a volumetric point cloud \cite{Qureshi2019} encoded with a multi-layer perpeceptron\,(MLP). We show that obstacles representation is critical to solve complex problems with rich workspace variations and propose to rely on a point cloud representation of obstacles coupled with PointNet \cite{Qi2017}. \citet{Qi2017} demonstrated the performance of PointNet to classify and segment point clouds, in this work we propose to use it to encode obstacles.

\section{Method}
\label{sec:prel}

\subsection{Overview of the method} 
\label{sec:pb_statement}

In this work we consider rigid robots with 2 to 6 degrees of freedom (DoF). Let $\mathcal{W} \subseteq \mathbb{R}^3$ be the workspace of the robot, containing a set of obstacles $\mathcal{V}$. We denote $\mathcal{C}_{free}$ the open set of configurations
where the robot does not collide with the obstacles $\mathcal{V}$ and 
$\mathcal{C}_{collision} = \mathcal{C} \setminus \mathcal{C}_{free}$.
Given a start configuration $q_0 \in \mathcal{C}_{free}$ and a goal configuration
$q_g \in \mathcal{C}_{free}$,  motion planning aims at finding
a \textit{collision-free} path which connects the start configuration to the goal configuration. 
A continuous function $\tau : [0, 1] \rightarrow \mathcal{C}$ is a solution if $\tau([0, 1])$ is a subset of $\mathcal{C}_{free}$, $\tau(0) = q_0$ and $\tau(1) = q_g$. A motion planning problem is thus defined by a start configuration, a goal configuration and a set of obstacles $\mathcal{V}$.

In this work, we represent obstacles $\mathcal{V}$ by sets of points (point clouds) and the corresponding normals sampled on the surface of obstacles. This representation is valid for arbitrary shapes and can be obtained with a depth sensor. We consider the robot as an embodied agent which senses surrounding obstacles with a panoramic camera (Fig.\ref{fig:overview}a). The point cloud, expressed in the robot local coordinate frame (Fig.\ref{fig:overview}b), is then processed with a PointNet architecture~\citep{Qi2017} to encode the obstacles (Fig.\ref{fig:overview}c).
The goal vector $q_g$ is expressed in the robot local coordinate frame and is concatenated to the PointNet obstacle encoding. This vector is processed by a MLP that generates actions, bringing the robot closer to the goal while avoiding obstacles (Fig.\ref{fig:overview}d). 
We learn the policy jointly with the PointNet encoding of the obstacles (Fig.\ref{fig:overview}c) in an end-to-end fashion with reinforcement learning. In the next two sections, we first give details on the obstacle representation and then describe policy learning.

\subsection{Obstacle representation for motion planning}
\label{sec:pb_representation}

We aim to learn a function that encodes the obstacles and the goal configuration as a vector enabling subsequent motion planning. While many parametric functions could be used as an encoder, we follow advances in visual recognition~\citep{Qi2017,krizhevsky2012imagenet,He_2016_CVPR} and define obstacle representations by a neural network learned jointly with the task of motion planning. We experimentally demonstrate the significant impact of the encoder on the performance of motion planning in Section~\ref{sec:experiments}.

In previous works \citep{Ichter2018, Jurgenson2019, Qureshi2019}, obstacles have been represented by  occupancy grids encoded with a MLP \citep{Ichter2018, Qureshi2019} or images encoded with a CNN \citep{Jurgenson2019} assuming global workspace knowledge. In our work we use points sampled on the surface of obstacles along with their oriented normals. Such measurements can be obtained using a depth sensor either placed on the robot or at other locations.
A set of obstacles $\mathcal{V}$ is represented by a finite set $S_{normals} = \{(x_i, n_i)\}_{i=1,...,N} \in \mathbb{R}^{N\times 2d}$ where $d=2, 3$ and the points $x_i$ and normals $n_i$ are expressed in the robot local coordinate frame. We denote by $\alpha_i$ the couple $(x_i, n_i)$. We define the goal $g$ as the displacement to reach the goal configuration from the current robot configuration.

To process a point cloud, we use a PointNet \citep{Qi2017} like network (Fig.\ref{fig:overview}c) reduced to its core layers for computation efficiency which we describe below. The idea of PointNet is to use a function which is symmetric with respect to the input to get a model invariant to input permutations. Based on this idea we build a network composed of two MLPs $u$ and $w$ with Exponential Linear Units (ELUs) activation \citep{Djork2015} as shown in Fig.\ref{fig:overview}c. The first MLP $u$ is shared across point cloud elements $\alpha_i$ (Fig.\ref{fig:overview}c blue block) and is used to generate local features of each element $u(\alpha_i)$. Then, a max operation is applied to obtain a global feature $v$ encoding the point cloud. The global feature $v$ is further concatenated with the goal $g$ (Fig.\ref{fig:overview}c green block) and passed through a second MLP $w$ (Fig.\ref{fig:overview}c orange block):
\begin{equation}
\begin{aligned}
v(\alpha_1, ..., \alpha_N) &= \underset{i=1,...,N}{\text{max}} \left[u(\alpha_1), ..., u(\alpha_N)\right] \\
f(\alpha_1, ..., \alpha_N, g) &= w(v(\alpha_1, ..., \alpha_N), g)
\end{aligned}
\end{equation}
The max operator is a max-pooling operation: given a list of feature vectors of size $N\times d$, it outputs a vector of size $d$ by taking the maximum over the $N$ points in each coordinate.

The PointNet encoding of obstacles $u$ is trained jointly with the policy $w$. For the training of $u$ and $w$ we compare imitation learning and reinforcement learning schemes as described in the next section.

\subsection{Learning policies for motion planning}
\label{sec:rl}

\comment{This section is devoted to briefly define reinforcement learning and then cast motion planning as a reinforcement learning problem. Reinforcement learning deals with optimal decision making in a Markov decision
process (MDP), defined by a tuple $(\mathcal{S}, \mathcal{O}, \mathcal{A}, p, r)$ \citep{Sutton2018}. $\mathcal{S}$ is the state space,  $\mathcal{A}$ is the action space, given two states $s_t$, $s_{t+1}$ and an action $a_t$, $p(s_{t+1}|s_t, a_t)$ is the transition function and $r(s_t, a_t)$ is the reward function. Finally, $\mathcal{O}$ is the observation space, $o_t = h(s_t)$ is an observable which informs either completely or partially about the system state, in the later case the process is a partially observable Markov decision process (POMDP). We will denote $\rho_\pi(s_t, a_t)$ as the state-action
distribution induced by a policy $\pi(a_t|s_t)$ and the transition function
$p(s_{t+1} | s_t, a_t)$. Maximum Entropy RL optimizes both the expected return and the entropy of the policy, the corresponding objective can be expressed as:

\begin{equation}
J(\pi) = \mathbb{E}_{(s_t, a_t) \sim \rho_\pi}\left[\sum_{t=0}^T r(s_t, a_t) -\alpha \log \pi(a_t|s_t)\right],
\end{equation}

where $R_t = \sum_t r(s_t, a_t)$ is the return and the entropy term incentivizes the policy to explore more widely \citep{SAC}.}

We cast the motion planning problem as a Markov decision
process (MDP) \cite{Sutton2018}.
The state space $\mathcal{S}$ is the configuration space of the robot
$\mathcal{C}$, $\mathcal{A}$ is the space of valid velocities at the current
configuration and $\mathcal{O}$ is a representation of the workspace along with the goal configuration $g$. Given a robot configuration $q \in \mathcal{C}_{free}$ and $v$ an admissible velocity vector, we denote $q(v)$ as the configuration reached by applying the velocity vector $v$ to the robot for a fixed time. As the robot can hit into obstacles, we consider $q_{free}(v)$ which returns the last collision free configuration
on the path connecting $q$ to $q(v)$. Then the dynamics $p$ is defined as 
$p(q, v) = q_{free}(v)$. 

We aim at learning policies to solve motion planning problems. For that purpose, we explore and compare policies trained with imitation learning (behavioral cloning) and reinforcement learning.
To train a policy with imitation learning \cite{Schaal2003}, we collect a dataset $\mathcal{D} = \{(o_t, a_t)\}$ of observation-action pairs along expert trajectories generated with Bi-RRT \cite{Kuffner2000} and follow the behavioral cloning approach \cite{Pomerleau1995}. Given a learnable policy $\pi$, we minimize the $L^2$ loss, $\mathcal{L}(\pi) = \|a_t-\pi(o_t)\|_2$, between the expert action $a_t$ and the policy applied to the expert observation $o_t$.

To train a policy with reinforcement learning \cite{Sutton2018}, we define a reward function as follows. Given a goal configuration $g \in \mathcal{C}$, we define $r_{velocity}(q, v) = -\|v\|_2$ and 
\begin{equation}
  r_{task}(q, v, g)=
  \begin{cases}
    r_{goal} ~\ \text{if} ~\ \|q_{free}(v)-g\| \leq \varepsilon, \\
    r_{free} ~\ \text{if} ~\ [q, q(v)] \subset \mathcal{C}_{free},  \\
    r_{collision} ~\ \text{else}.
  \end{cases}
\end{equation}
with $r_{goal} > 0$, $r_{free} < 0$ and $r_{collision} < 0$. The reward function is then defined as  $r(q, v, g)~=~r_{velocity}(q, v)~+~r_{task}(q, v, g)$. $r_{task}$ rewards actions reaching $g$ and penalizes actions which lead to a collision. Given two collision-free paths leading to the goal, the total reward $r(q,v,g)$ is highest for the shortest path. 
Maximizing the reward enables the policy path to be collision free and as short as possible.
Note that the dynamics $p$ depends only on the robot and the workspace, and the reward function $r$ depends additionally on the goal configuration to be reached. An episode is terminated when reaching the goal or the maximum number of steps.

The reward $r_{task}$ defined above is sparse with respect to the goal: it is only positive if the agent reaches the goal during one episode, which may have a low probability in challenging environments. Hindsight Experience Replay (HER)~\citep{Andrychowicz2017} is a technique to improve the sample efficiency of off-policy RL algorithms in the sparse and goal-conditioned setting which we use extensively in this work. After collecting one rollout $s_0, ..., s_T$ which may or may not have reached the goal $g$, it consists in sampling one of the states as the new goal $g'$ for this rollout. The rollout may not have reached $g$ but in hindsight, it can be considered as a successful rollout to reach $g'$. HER can be seen as a form of implicit curriculum learning which accelerates the learning of goal-conditioned policies.

\vspace{-0.2cm}
\section{Experimental Results}
\label{sec:experiments}
\vspace{-0.2cm}

Below we describe our experimental setup and implementation details in Sections~\ref{sec:envs} and~\ref{sec:implem_details}. Section~\ref{sec:exp_representation} evaluates alternative obstacle representations while Sections~\ref{sec:exp_local_global} and \ref{sec:exp_slot} compare our approach to the state of the art in challenging environments.

\vspace{-0.2cm}
\subsection{Environments}
\label{sec:envs}

We evaluate our method in a number of different environments, namely, 2D and 3D environments used in~\cite{Ichter2018,Qureshi2019}, our own 3D environments and an environment based on a classic motion planning problem~\cite{Latombe1998} where a S-shape with 6\,DoF should go through a thin slot. We consider rigid body robots which are either a 2D/3D sphere with 2/3\,DoF or a S-shape body with 6\,DoF.
We use distinct workspaces for training and evaluation so that policies are evaluated on workspaces unseen during training. We evaluate the success rate of a policy over 400 rollouts. 
At the end of a rollout, the environment is reset: a new random workspace is sampled along with a start and goal configuration. A rollout is considered successful if it reaches a configuration near the goal defined by an epsilon neighborhood before the maximum number of steps is reached. 
We describe details for each of our environments below.

\textit{2D-Narrow \cite{Ichter2018}:} we generate 2D environments from \citet{Ichter2018} using publicly available code \cite{IchterCVAE}. The environment contains a sphere robot navigating in 2D workspaces composed of 3 randomly generated narrow gaps as shown in Fig.\ref{fig:2d_representations} and random start and goal configurations. We set the maximum number of policy steps to 50.\\
\textit{3D-Qureshi \cite{Qureshi2019}:} 3D workspaces with a sphere robot from \citet{Qureshi2019} contain axis-aligned boxes with fixed center and varying sizes. We used open-source code from \citet{QureshiMPNet} to generate the workspaces.\\
\textit{3D-Boxes:} our environment composed of 3 to 10 static boxes generated with random sizes, positions and orientations as illustrated in Fig.\ref{fig:3d_envs}(a). The maximum number of steps is set to 80.\\
\textit{3D-Synthethic:} a variant of 3D-Boxes composed of unseen synthetic obstacles such as capsules, cylinders and spheres instead of boxes as illustrated in Fig.\ref{fig:3d_envs}(b).\\
\textit{3D-YCB:} a variant of 3D-Boxes composed of real objects from the YCB dataset \cite{Calli2015} recorded with a RGB-D camera, see Fig.\ref{fig:3d_envs}(c).\\ \textit{3D-Dynamic:} a variant of 3D-Boxes with dynamic obstacles moving in real time. For each box two placements are sampled uniformly and the current box placement is interpolated between the two.\\
\textit{Slot:} S-shaped rigid 6\,DoF robot that should pass through a thin slit of varying width 2-8 times wider than the smallest robot link. This is a classic problem in motion planning \cite{Latombe1998} illustrated in Fig.\ref{fig:3d_envs}(d).

\vspace{-0.2cm}
\subsection{Implementation details}
\label{sec:implem_details}
\vspace{-0.2cm}

Below we describe implementation details for our own and other methods used for comparison.
To train policies with reinforcement learning, we use Soft Actor Critic (SAC) \cite{SAC} with automatic entropy tuning, a learning rate of $3.10^{-4}$ with Adam optimizer \cite{Kingma2014}, a batch size of $256$ and a replay buffer of $10^6$ steps combined with Hindsight Experience Replay (HER) \cite{Andrychowicz2017} with $80\%$ of the trajectories goal relabeled, as in the original papers. We train a policy on $2.10^6$ environment steps before reporting results. We use the open-source implementation of \citet{Vitchyr2020}. The policy $\pi$ and the $Q$-function are implemented as separate networks and trained jointly with alternate optimization following the actor-critic scheme. We use a PointNet based policy with $3$ hidden layers of size $256$ for the point encoder $u$ and the global feature network $w$ respectively as shown in Fig.\ref{fig:overview}c. The $Q$-function has a similar structure and the action is concatenated to each point of the point cloud. \vspace{-.1cm}

For the comparison of representations in Table~\ref{tab:2d_representation} we use a CNN policy with $64\times 64$ image inputs composed of 3 convolutional layers with 32 filters of size $3 \times 3$ followed by a max-pooling operator and $3$ hidden layers of size $256$. The current configuration and goal are concatenated after max-pooling. For the $Q$-function, the action is also concatenated after max-pooling. The MLP policy processing point clouds is composed of $3$ hidden layers of size $256$ and the goal is concatenated to the list of points. For the $Q$-function, the action is also concatenated to the input. \vspace{-.1cm}

To train policies with imitation learning, we use a learning rate of $10^{-3}$ with Adam optimizer \cite{Kingma2014} and a batch size of $256$. For Bi-RRT, once a solution is found, we shorten its length by randomly sampling two points along the solution, if the shortcut made out of these two points is collision free we modify the solution to include it. The maximal size of an edge in RRT corresponds to the maximal size of a policy step for RL or BC, in this way computing collision checking on an edge has the same cost for both. \vspace{-.1cm}

For \citet{Ichter2018}, we used the code provided in \cite{IchterCVAE} along with the dataset to train the conditional variational auto-encoder. Once trained we combined the learned sampling with Bi-RRT to report the results.
For \citet{Qureshi2019}, we adjust the implementation provided by \citet{QureshiMPNet} to our environments. We followed the training procedure described in \cite{Qureshi2019}. For Qureshi neural replanning (NR), we run neural planning for $50$ steps then use neural replanning recursively for $10$ iterations each comprising $50$ steps maximum. For Qureshi hybrid replanning (HR), if neural replanning fails to find a solution after $10$ iterations, Bi-RRT is used to find a path between states where there is still a collision.
For \citet{Jurgenson2019}, we adapted the open-source implementation \cite{TomJurDDPGMP} to run on our environment. For training, we use the same parameters as \cite{Jurgenson2019}, image based workspace representation and reward definition. We follow their DDPG-MP method. The training dataset consists of $10^4$ workspaces. For pre-training, we use $400$ random steps per workspace and for training, we use Bi-RRT generated $10$ expert trajectories per workspace.
Finally, the rigid bodies and obstacles are modeled using Pinocchio \cite{pinocchioweb, carpentier2019pinocchio} and collision checks are computed by FCL~\cite{pan2012fcl}.
\vspace{-.4cm}

\subsection{Comparison of obstacle representations and learning approaches}
\label{sec:exp_representation}

\begin{figure*}
  \centering
    \includegraphics[width=0.9\textwidth]{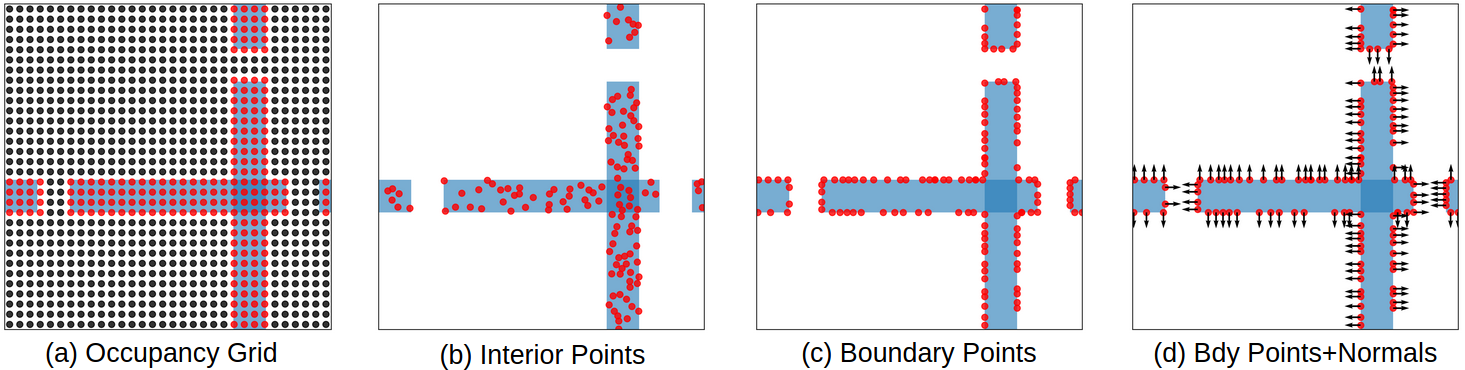}
  \caption{Illustration of different obstacles representations for 2D-Narrow. See Table~\ref{tab:2d_representation} for results.}
  \label{fig:2d_representations}
\vspace{-0.1cm}
\end{figure*}

In this section we compare different policies on the 2D-Narrow environment \cite{Ichter2018}. Policies are trained with Behavioral Cloning (BC) or Reinforcement Learning (RL) using different obstacles representations presented in Fig.\ref{fig:2d_representations}. The occupancy grid is a $64\times 64$ image (Fig.\,\ref{fig:2d_representations}a), the point clouds are composed of 128 points either sampled on the interior of obstacles (Fig.\,\ref{fig:2d_representations}b) or at their boundary (Fig.\,\ref{fig:2d_representations}c, \ref{fig:2d_representations}d). 128 points is a good trade-off between speed and accuracy as adding more points did not improve our results. In Table \ref{tab:2d_representation} we report the success rate of RL policies after 1000 epochs which corresponds to $10^6$ interactions with the environment, a stage at which the policies performance have plateaued. For BC, we collect a dataset of solutions using Bi-RRT, containing $10^6$ steps in total and train for 200 epochs. This allows for a fair comparison of BC and RL  trained on the same dataset size. A simple baseline connecting the start and the goal by a straight line has success rate of 45.5\%.

\begin{table}
\small
\setlength\tabcolsep{2.7pt} 
\centering
\begin{tabular}{ ccc } 
 \toprule
  & Behavioral Cloning & Reinforcement Learning \\
  \midrule
  64$\times$64 Image - CNN & 64.5\% & 44.5\% \\
  Interior points - MLP & 42.0\%  & 29.0\% \\
  Interior points - PointNet & 76.0\% & 83.0\% \\
  Boundary points - PointNet & 85.0\% & 93.8\% \\
  Boundary points and normals - PointNet & 86.5\% & \textbf{99.5\%} \\ 
  \bottomrule
\end{tabular}
\caption{Comparison of different obstacles representations and policies training methods on the 2D-Narrow environment.}
\mbox{}
\label{tab:2d_representation}
\vspace{-0.9cm}
\end{table}

Table \ref{tab:2d_representation} shows that the choice of obstacles representation greatly impacts the policy success rate both for BC and RL. Using a representation based on an occupancy grid encoded with a CNN yields poor performance for both BC and RL. Similarly, if interior points are encoded with a MLP the performance is low for both BC and RL. In both case results with RL are below BC. For BC, the training set is composed of expert demonstrations which is fixed for every representation. In contrast, as the RL training set is generated by the learning policy, the obstacles representation impacts the quality of the training set. Encoding interior points with PointNet results in a significant gain, +34\% for BC and +54\% for RL. Using PointNet in combination with boundary points and normals increases the performance by +10\% for BC and +16\% for RL. We can also observe that normals improve the performance over boundary points alone. Furthermore, with a more stable obstacle encoding, RL outperforms BC by a margin, +13\% in the case of boundary points and normals. 

\begin{figure*}[htbp]
  \centering
    \includegraphics[width=0.9\textwidth]{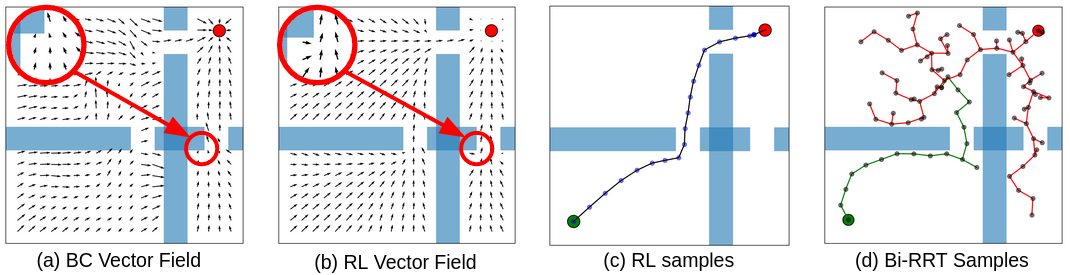}
  \caption{(a) Vector field of a policy trained with behavioral cloning. (b) Vector field of a policy trained with reinforcement learning. (c) RL path samples. (d) RRT samples generated to find a path. }
  \label{fig:samples}
\end{figure*}

The difference between policies trained with BC or RL is illustrated in Fig.\ref{fig:samples}(a,b). For each policy, a vector field has been generated by using a grid covering the environment and computing the policy output at each point given a fixed goal plotted in red. We observe that policies trained with BC can fail close to edges of obstacles.  This is a typical problem of imitation learning~\cite{Ross2011} which is limited to perfect, expert trajectories in the training set and does not observe failure cases. In contract, RL explores the environment during training and generates actions pointing away from obstacles as it has been trained explicitly to avoid collisions.

We also evaluate alternative obstacle representations on a collision classification task in the 2D-Narrow environment (see Fig.\ref{fig:2d_representations}). Given a configuration and an action, the goal is to predict if the robot configuration after executing the action is colliding with any obstacle or not. We compare classifiers trained on top of (a) 64x64 images encoded with CNNs, (b) interior points, (c) boundary points and (d) boundary points with normals, where (b), (c) and (d) are encoded with PointNet. 
The resulting accuracies for collision prediction are: (a) 80\%, (b) 94\%, (c) 97\% and (d) 98.5\%. Consistently with results in Table \ref{tab:2d_representation}, we observe that (i)~PointNet representations outperform occupancy grid with CNN and (ii)~boundary points with normals encoding give best results. Further analysis of failure cases has shown inaccuracy of CNN collision predictions at locations near the boundaries of obstacles and in regions with narrow passages.

\begin{table}
\small
\setlength\tabcolsep{2.7pt} 
\centering
\comment{\begin{tabular}{ ccccc } 
 \toprule
  & Fixed Workspace & Random Workspaces & Nodes & Path Length \\
  \midrule
  \comment{Bi-RRT~\cite{Kuffner2000} & 100\% & 100\% & 265 & 1.38  \\}
  Bi-RRT~\cite{Kuffner2000} & 100\% & 100\% & 358 & 0.65\\
  Ichter $\lambda=0.5$ \cite{Ichter2019} & 100\% & 100\% & 232 & 1.30 \\
  Ichter $\lambda=0.9$ \cite{Ichter2019} & 100\% & 100\% & 207 & 1.22 \\
  Qureshi (NR) \cite{Qureshi2019} & 99.5\% & 68.0\% & 102 & 0.52 \\
  Qureshi (HR) \cite{Qureshi2019} & 99.5\% & 95.0\% & 950 & 0.69 \\
  Jurgenson \cite{Jurgenson2019} & 74.0\% & 47.0\% & 12 & 0.56\\ 
  Us & 99.5\% & 99.5\% & 10 & 0.63  \\
  \bottomrule
\end{tabular}}
\begin{tabular}{ ccccc } 
 \toprule
  & Success Rate & Nodes & Path Length \\
  \midrule
  \comment{Bi-RRT~\cite{Kuffner2000} & 100\% & 265 & 1.38  \\}
  Bi-RRT~\cite{Kuffner2000} & 100\% & 358 & 0.65\\
  Ichter $\lambda=0.5$ \cite{Ichter2018} & 100\% & 232 & 1.30 \\
  Ichter $\lambda=0.9$ \cite{Ichter2018} & 100\% & 207 & 1.22 \\
  Qureshi (NR) \cite{Qureshi2019} & 68.0\% & 102 & 0.52 \\
  Qureshi (HR) \cite{Qureshi2019} & 95.0\% & 950 & 0.69 \\
  Jurgenson \cite{Jurgenson2019} & 47.0\% & 12 & 0.56\\ 
  Us & 99.5\% & 10 & 0.63  \\
  \bottomrule
\end{tabular}
\caption{Comparison to the state of the art on the 2D-Narrow environment.}
\mbox{}
\label{tab:2d_sota}
\vspace{-0.9cm}
\end{table}

We compare our approach to state-of-the-art neural motion planning approaches~\cite{Ichter2018, Jurgenson2019, Qureshi2019} and Bi-RRT~\cite{Kuffner2000} in Table \ref{tab:2d_sota}. We compare in terms of success rate, number of configurations (nodes) explored before finding a successful path and in terms of length of the found solution. We chose to compare the number of configurations explored to find a solution because connecting two configurations requires to perform collision checking which represents 95\% of time spent by motion planning algorithms \cite{Ratliff2009}. Bi-RRT achieves a success rate of 100\% as a solution is found if the algorithm is run long enough but it requires 35 times more nodes than RL to find a solution and shorten the path. \citet{Ichter2018}, which is also based on RRT, requires 20 times more nodes than RL to find a solution and yields longer solutions overall. The neural replanning (NR) and hybrid replanning (HR) approaches of \citet{Qureshi2019} allow to find short paths at the price of extensive use of collision checking, which is limiting in scenarios with time constraints. \citet{Jurgenson2019} uses an image representation of obstacles, yielding a low success rate which correlates with the results of Table \ref{tab:2d_representation}, it mostly solves problems with straight solutions which explains the short path length.

 In Fig.\ref{fig:samples}(c,d) we illustrate the number of nodes required to find a path.  While RL outputs short paths leading directly to the goal (Fig.\ref{fig:samples}c), Bi-RRT explores the space in several directions before finding a suboptimal path which then needs to be shortened (Fig.\ref{fig:samples}d).

\vspace{-0.2cm}
\subsection{Towards a realistic setup: 3D environments with local observability}
\label{sec:exp_local_global}

We compare agents trained with either global observations, where points are sampled for all obstacles surface as in Section \ref{sec:exp_representation}, or local observations where obstacles are only observed locally, through a camera which is closer to a realistic setup. For the local observation, we consider agents equipped with a panoramic camera observing the local geometry (Fig.\ref{fig:overview}b). To model such agents, we use ray tracing and cast rays from the center of the robot in every direction by uniformly sampling points on the sphere. Each ray hitting an obstacle provides a point with its normal and the agent observation consists of the point cloud formed by the union of these points. For local observations (\texttt{Local}), we use a point cloud of 64 points, corresponding to a density of 60 points per meter squared. For global observations (\texttt{Global}), we use 256 points sampled uniformly on the obstacles surface which corresponds to the same point density as \texttt{Local}. The two observations thus have the same geometry resolution.

\begin{figure*}[htbp]
  \centering
    \includegraphics[width=\textwidth]{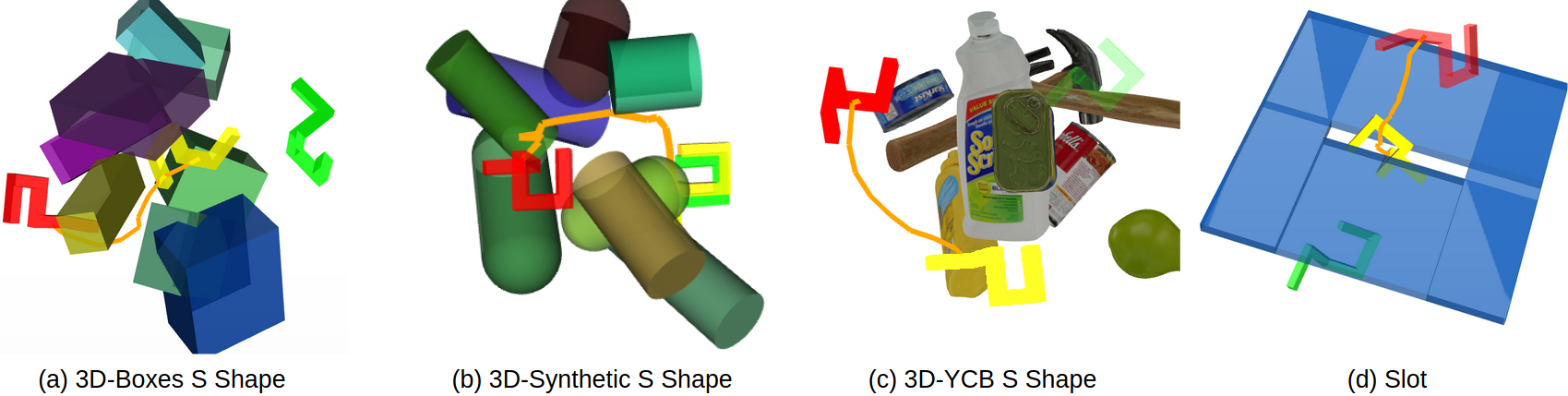}
  \caption{3D environments with S-shaped robot composed of: (a) boxes obstacles, (b) sphere, cylinder and cone obstacles, (c) YCB dataset \cite{Calli2015} obstacles, (d) a thin slot. We plot the start configuration in red, the current configuration in yellow and the goal configuration in green.}
  \label{fig:3d_envs}
\vspace{-0.3cm}
\end{figure*}

We consider 3D environments and compare agents controlling either a sphere robot with 3DoF or a S-shaped robot with 6DoF as shown in Fig.\ref{fig:3d_envs} and report results in Table \ref{tab:3d_local_global}. All the policies are trained on the 3D-Boxes environment (Fig.\ref{fig:3d_envs}a) exclusively. The S-shape problem is harder to solve than the sphere one but our approach still yields good results with a performance of $97\%$ on 3D-Boxes for the local observation. Overall \texttt{Local} and \texttt{Global} policies yield similar performances which shows that our approach still works on harder problems where only local knowledge of obstacles is available. When tested on 3D-Boxes, \texttt{Local} yields better results than \texttt{Global} while the generalization performance are better for \texttt{Global} when tested on unseen environments (Fig.\ref{fig:3d_envs}b,c).
While solely trained on problems with static obstacles from 3D Boxes (Fig.\ref{fig:3d_envs}a), the policies generalize to unseen scenes containing new set of synthetic obstacles (3D-Synthetic) and real obstacles recorded with depth sensor from the YCB dataset (3D-YCB). The policies trained with our approach also solve challenging scenarios with dynamic obstacles moving in real time (3D-Dynamic). This highlights the advantage of using a policy which directly computes the next action instead of RRT-based approaches relying on offline path planning \citep{Kuffner2000, Qureshi2019, Ichter2019}.

\begin{table}[htbp]
\small
\setlength\tabcolsep{2.7pt} 
\centering
\begin{tabular}{ ccccc } 
 \toprule
  & 3D-Boxes & 3D-Synthetic & 3D-YCB & 3D-Dynamic \\
  \midrule
  Sphere - \texttt{Global} & 95.8\% & 91.3\% & 91.2\% & 94.7\% \\
  S Shape - \texttt{Global} & 89.0\% & 85.3\% & 88.7\% & 86.0\% \\
  \midrule
  Sphere - \texttt{Local} & 99.3\% & 87.0\% & 87.7\% & 95.8\% \\
  S Shape - \texttt{Local} & 97.0\% & 87.3\% & 96.0\% & 79.7\% \\
  \bottomrule
\end{tabular}   
\caption{Comparison of sphere and S-shaped agents trained with global or local obstacle observation.}
\mbox{}
\label{tab:3d_local_global}
\vspace{-0.6cm}
\end{table}
\begin{table}[htbp]
\small
\setlength\tabcolsep{2.7pt} 
\centering
\begin{tabular}{ cccc } 
 \toprule
  & 3D-Qureshi \cite{Qureshi2019} & 3D-Boxes & 3D-Synthetic \\
  \midrule
  Qureshi (NR) \cite{Qureshi2019} - MLP & 96.0\% & 84.0\% & 81.0\% \\
  Qureshi (HR) \cite{Qureshi2019} - MLP & 99.5\% & 88.0\% & 84.5\% \\
  Sphere - \texttt{Global} & \textbf{100\%} & \textbf{95.8\%} & \textbf{91.3\%} \\
  \bottomrule
\end{tabular}
\caption{Comparison of our approach to Qureshi \cite{Qureshi2019} on Qureshi's environment and our environment}
\mbox{}
\label{tab:3d_qureshi}
\vspace{-0.6cm}
\end{table}

We also compare our approach to \citet{Qureshi2019}. For a fair comparison we use global obstacles representation for our approach as \cite{Qureshi2019} uses full obstacles knowledge and report results in Table \ref{tab:3d_qureshi}. On 3D-Qureshi, we show that our approach successfully solves the proposed problems. On 3D-Boxes and 3D-Synthetic we show that our approach has better generalization abilities while only needing $20$ nodes on average to solve problems where \cite{Qureshi2019} requires more than $400$ nodes.
\vspace{-0.2cm}

\subsection{S-shape motion planning}
\label{sec:exp_slot}
\vspace{-0.2cm}

We consider a challenging problem in motion planning composed of a S-shaped robot and a thin gate it has to go through, introduced by \citet{Latombe1998} and shown in Fig.\ref{fig:3d_envs}(d). The width of the gate determines the difficulty of the problem. We consider problems with a gate of random width, sampled to be 2 times to 8 times wider than the smallest dimension of each robot link, as a comparison, the gate of \citep{Latombe1998} is 2.5 times wider. 
We compare the performance of an agent trained with \texttt{Local} observations to Bi-RRT with an allocated budget of 50000 nodes for each problem. The learned policy has a success rate of 97.7\% whereas Bi-RRT has a success rate of 62.5\% on average when tested on 400 planning problems. In contrast with experiments of Section \ref{sec:exp_representation}, Bi-RRT does not succeed to solve every problem with the allocated nodes. While our trained policy provides solutions in real-time composed of 40 nodes on average, the computational burden of RRT is increasing significantly as the number of explored configurations increases which is typically the case for this environment where many nodes need to be expanded to find a solution. We have also noted that our policy adapts its behavior to minimize the path length according to the slot size. Indeed, when the slot is thin, e.g. 2 times wider than the smallest robot link, the motions are quite constrained, the policy inserts a link by translating the S-shape, rotates by 90 degrees and translates again which was also the only solution found when using RRT. When the slot is wider we observe that the policy uses the wider space provided by the diagonal of the slot to reduce overall motion. 

\vspace{-0.4cm}
\section{Conclusion}
\label{sec:conclusion}
\vspace{-0.2cm}

This paper introduces a new framework for neural motion planning. Obstacles are represented by point clouds and encoded by a PointNet architecture. PointNet encoding and motion policy are trained jointly with either behavioral cloning or reinforcement learning. We show that PointNet encoding outperforms state-of-the-art representations based on image-based CNNs and latent representations. Future work will address rigid robots with multiple links as for example robotic arms performing manipulation tasks in cluttered environments directly captured by a camera.

{\small
{\bfseries Acknowledgments.} We thank Lo\"ic Est\`eve for the helpful discussions. This work was partially supported by the HPC resources from GENCI-IDRIS  (Grant 20XX-AD011011163), Louis Vuitton ENS Chair on Artificial Intelligence, and the French government under management of Agence Nationale de la Recherche as part of the ”Investissements d’avenir” program, reference ANR-19-P3IA-0001 (PRAIRIE 3IA Institute). }

\small
\bibliography{references}

\begin{thebibliography}{41}
\providecommand{\natexlab}[1]{#1}
\providecommand{\url}[1]{\texttt{#1}}
\expandafter\ifx\csname urlstyle\endcsname\relax
  \providecommand{\doi}[1]{doi: #1}\else
  \providecommand{\doi}{doi: \begingroup \urlstyle{rm}\Url}\fi

\bibitem[Qi et~al.(2017)Qi, Su, Mo, and Guibas]{Qi2017}
C.~R. Qi, H.~Su, K.~Mo, and L.~J. Guibas.
\newblock {PointNet: Deep learning on point sets for 3D classification and
  segmentation}.
\newblock In \emph{Proceedings - 30th IEEE Conference on Computer Vision and
  Pattern Recognition, CVPR 2017}, 2017.
\newblock
\newblock ISBN 9781538604571.

\bibitem[Latombe(1991)]{Latombe1991}
J.-C. Latombe.
\newblock \emph{Robot Motion Planning}.
\newblock Kluwer Academic Publishers, USA, 1991.
\newblock ISBN 079239206X.

\bibitem[LaValle(2006)]{Lavalle2006}
S.~M. LaValle.
\newblock \emph{Planning Algorithms}.
\newblock Cambridge University Press, USA, 2006.
\newblock ISBN 0521862051.

\bibitem[Fox et~al.(1997)Fox, Burgard, and Thrun]{Fox1997}
D.~Fox, W.~Burgard, and S.~Thrun.
\newblock The dynamic window approach to collision avoidance.
\newblock \emph{{IEEE} Robotics Autom. Mag.}, \penalty0 (1):\penalty0 23--33,
\newblock
\newblock 1997.

\bibitem[{Laumond}(2006)]{Laumond2006}
J.-P. {Laumond}.
\newblock Kineo cam: a success story of motion planning algorithms.
\newblock \emph{IEEE Robotics Automation Magazine}, \penalty0 (2):\penalty0
  90--93, 2006.

\bibitem[Harada et~al.(2014)Harada, Yoshida, and Yokoi]{Harada2014}
K.~Harada, E.~Yoshida, and K.~Yokoi.
\newblock \emph{Motion Planning for Humanoid Robots}.
\newblock Springer Publishing Company, Incorporated, 2014.
\newblock ISBN 1447157052.

\bibitem[Jr. and LaValle(2000)]{Kuffner2000}
J.~J.~K. Jr. and S.~M. LaValle.
\newblock Rrt-connect: An efficient approach to single-query path planning.
\newblock In \emph{Proceedings of the 2000 {IEEE} International Conference on
  Robotics and Automation, {ICRA} 2000, April 24-28, 2000, San Francisco, CA,
  {USA}}, pages 995--1001. {IEEE},
\newblock
\newblock 2000.

\bibitem[Kavraki et~al.(1996)Kavraki, Svestka, Latombe, and
  Overmars]{Kavraki1996}
L.~E. Kavraki, P.~Svestka, J.~Latombe, and M.~H. Overmars.
\newblock Probabilistic roadmaps for path planning in high-dimensional
  configuration spaces.
\newblock \emph{{IEEE} Trans. Robotics Autom.}, \penalty0 (4):\penalty0
  566--580,
\newblock
\newblock 1996.

\bibitem[Schwartz et~al.(1986)Schwartz, Sharir, and Hopcroft]{Sharir1986}
J.~T. Schwartz, M.~Sharir, and J.~E. Hopcroft, editors.
\newblock \emph{Planning, Geometry, and Complexity of Robot Motion}.
\newblock Ablex Publishing Corp., USA, 1986.
\newblock ISBN 0893913618.

\bibitem[Glasius et~al.(1995)Glasius, Komoda, and Gielen]{Glasius1995}
R.~Glasius, A.~Komoda, and S.~C. Gielen.
\newblock {Neural Network Dynamics for Path Planning and Obstacle Avoidance}.
\newblock \emph{Neural Networks}, 1995.
\newblock
\newblock ISSN 08936080.

\bibitem[Yang and Meng(2000)]{Yang2000}
S.~X. Yang and M.~Meng.
\newblock {An efficient neural network approach to dynamic robot motion
  planning}.
\newblock \emph{Neural Networks}, 2000.
\newblock
\newblock ISSN 08936080.

\bibitem[Pfeiffer et~al.(2017)Pfeiffer, Schaeuble, Nieto, Siegwart, and
  Cadena]{Pfeiffer2017}
M.~Pfeiffer, M.~Schaeuble, J.~Nieto, R.~Siegwart, and C.~Cadena.
\newblock {From perception to decision: A data-driven approach to end-to-end
  motion planning for autonomous ground robots}.
\newblock In \emph{Proceedings - IEEE International Conference on Robotics and
  Automation}, 2017.
\newblock
\newblock ISBN 9781509046331.

\bibitem[Ichter et~al.(2018)Ichter, Harrison, and Pavone]{Ichter2018}
B.~Ichter, J.~Harrison, and M.~Pavone.
\newblock {Learning Sampling Distributions for Robot Motion Planning}.
\newblock In \emph{Proceedings - IEEE International Conference on Robotics and
  Automation}, 2018.
\newblock
\newblock ISBN 9781538630815.

\bibitem[Jurgenson and Tamar(2019)]{Jurgenson2019}
T.~Jurgenson and A.~Tamar.
\newblock Harnessing reinforcement learning for neural motion planning.
\newblock \emph{RSS},
\newblock 2019.

\bibitem[Qureshi et~al.(2019)Qureshi, Simeonov, Bency, and Yip]{Qureshi2019}
A.~H. Qureshi, A.~Simeonov, M.~J. Bency, and M.~C. Yip.
\newblock {Motion planning networks}.
\newblock In \emph{Proceedings - IEEE International Conference on Robotics and
  Automation}, 2019.
\newblock
\newblock ISBN 9781538660263.

\bibitem[Krizhevsky et~al.(2012)Krizhevsky, Sutskever, and
  Hinton]{krizhevsky2012imagenet}
A.~Krizhevsky, I.~Sutskever, and G.~E. Hinton.
\newblock Imagenet classification with deep convolutional neural networks.
\newblock In \emph{Advances in neural information processing systems (NIPS)},
  pages 1097--1105, 2012.

\bibitem[He et~al.(2016)He, Zhang, Ren, and Sun]{He_2016_CVPR}
K.~He, X.~Zhang, S.~Ren, and J.~Sun.
\newblock Deep residual learning for image recognition.
\newblock In \emph{Proceedings of the IEEE Conference on Computer Vision and
  Pattern Recognition (CVPR)}, 2016.

\bibitem[Haarnoja et~al.(2018)Haarnoja, Zhou, Hartikainen, Tucker, Ha, Tan,
  Kumar, Zhu, Gupta, Abbeel, and Levine]{SAC}
T.~Haarnoja, A.~Zhou, K.~Hartikainen, G.~Tucker, S.~Ha, J.~Tan, V.~Kumar,
  H.~Zhu, A.~Gupta, P.~Abbeel, and S.~Levine.
\newblock Soft actor-critic algorithms and applications.
\newblock \emph{CoRR},
\newblock 2018.

\bibitem[nmp(2020)]{nmpwebpage}
Learning obstacle representations for neural motion planning, project webpage.
\newblock \url{https://www.di.ens.fr/willow/research/nmp_repr/}, 2020.

\bibitem[Latombe()]{Latombe1998}
L.~E. K. J.-C. Latombe.
\newblock Probabilistic roadmaps for robot path planning.

\bibitem[Jetchev and Toussaint(2010)]{Jetchev2010}
N.~Jetchev and M.~Toussaint.
\newblock {Trajectory prediction in cluttered voxel environments}.
\newblock In \emph{Proceedings - IEEE International Conference on Robotics and
  Automation}, 2010.
\newblock
\newblock ISBN 9781424450381.

\bibitem[Lien and Lu(2010)]{Lien2010}
J.~M. Lien and Y.~Lu.
\newblock {Planning motion in environments with similar obstacles}.
\newblock In \emph{Robotics: Science and Systems}, 2010.
\newblock
\newblock ISBN 9780262514637.

\bibitem[Branicky et~al.(2008)Branicky, Knepper, and Kuffner]{Branicky2008}
M.~S. Branicky, R.~A. Knepper, and J.~J. Kuffner.
\newblock {Path and trajectory diversity: Theory and algorithms}.
\newblock In \emph{Proceedings - IEEE International Conference on Robotics and
  Automation}, 2008.
\newblock
\newblock ISBN 9781424416479.

\bibitem[Martin et~al.(2007)Martin, Wright, and Sheppard]{Martin2007}
S.~R. Martin, S.~E. Wright, and J.~W. Sheppard.
\newblock {Offline and online evolutionary bi-directional RRT algorithms for
  efficient re-planning in dynamic environments}.
\newblock In \emph{Proceedings of the 3rd IEEE International Conference on
  Automation Science and Engineering, IEEE CASE 2007}, 2007.
\newblock
\newblock ISBN 1424411548.

\bibitem[Clevert et~al.(2016)Clevert, Unterthiner, and Hochreiter]{Djork2015}
D.~Clevert, T.~Unterthiner, and S.~Hochreiter.
\newblock Fast and accurate deep network learning by exponential linear units
  (elus).
\newblock In Y.~Bengio and Y.~LeCun, editors, \emph{4th International
  Conference on Learning Representations, {ICLR} 2016, San Juan, Puerto Rico,
  May 2-4, 2016, Conference Track Proceedings},
\newblock 2016.

\bibitem[Sutton and Barto(2018)]{Sutton2018}
R.~S. Sutton and A.~G. Barto.
\newblock \emph{Reinforcement Learning: An Introduction}.
\newblock A Bradford Book, Cambridge, MA, USA, 2018.
\newblock ISBN 0262039249.

\bibitem[Schaal et~al.(2003)Schaal, Ijspeert, and Billard]{Schaal2003}
S.~Schaal, A.~Ijspeert, and A.~Billard.
\newblock Computational approaches to motor learning by imitation.
\newblock \penalty0 (1431):\penalty0 537--547,
\newblock 2003.
\newblock clmc.

\bibitem[Pomerleau(1989)]{Pomerleau1995}
D.~A. Pomerleau.
\newblock Alvinn: An autonomous land vehicle in a neural network.
\newblock In \emph{Advances in Neural Information Processing Systems 1}, San
  Francisco, CA, USA, 1989. Morgan Kaufmann Publishers Inc.
\newblock ISBN 1558600159.

\bibitem[Andrychowicz et~al.(2017)Andrychowicz, Wolski, Ray, Schneider, Fong,
  Welinder, McGrew, Tobin, Abbeel, and Zaremba]{Andrychowicz2017}
M.~Andrychowicz, F.~Wolski, A.~Ray, J.~Schneider, R.~Fong, P.~Welinder,
  B.~McGrew, J.~Tobin, P.~Abbeel, and W.~Zaremba.
\newblock {Hindsight experience replay}.
\newblock In \emph{Advances in Neural Information Processing Systems}, 2017.

\bibitem[Pong(2020)]{IchterCVAE}
V.~Pong.
\newblock Learning sampling distributions.
\newblock https://github.com/StanfordASL/LearnedSamplingDistribution, 2020.

\bibitem[Qureshi(2020)]{QureshiMPNet}
A.~Qureshi.
\newblock Implementation of mpnet: Motion planning networks.
\newblock \url{https://github.com/ahq1993/MPNet}, 2020.

\bibitem[{Calli} et~al.(2015){Calli}, {Singh}, {Walsman}, {Srinivasa},
  {Abbeel}, and {Dollar}]{Calli2015}
B.~{Calli}, A.~{Singh}, A.~{Walsman}, S.~{Srinivasa}, P.~{Abbeel}, and A.~M.
  {Dollar}.
\newblock The ycb object and model set: Towards common benchmarks for
  manipulation research, 2015.

\bibitem[Kingma and Ba(2015)]{Kingma2014}
D.~P. Kingma and J.~Ba.
\newblock Adam: {A} method for stochastic optimization.
\newblock In Y.~Bengio and Y.~LeCun, editors, \emph{3rd International
  Conference on Learning Representations, {ICLR} 2015, San Diego, CA, USA, May
  7-9, 2015, Conference Track Proceedings},
\newblock 2015.

\bibitem[Pong(2020)]{Vitchyr2020}
V.~Pong.
\newblock Reinforcement learning framework and algorithms implemented in
  pytorch.
\newblock \url{https://github.com/vitchyr/rlkit}, 2020.

\bibitem[Jurgenson(2020)]{TomJurDDPGMP}
T.~Jurgenson.
\newblock Implementation of ddpg-mp.
\newblock \url{https://github.com/tomjur/ModelBasedDDPG}, 2020.

\bibitem[Carpentier et~al.(2019{\natexlab{a}})Carpentier, Valenza, Mansard,
  et~al.]{pinocchioweb}
J.~Carpentier, F.~Valenza, N.~Mansard, et~al.
\newblock Pinocchio: fast forward and inverse dynamics for poly-articulated
  systems.
\newblock https://stack-of-tasks.github.io/pinocchio, 2019{\natexlab{a}}.

\bibitem[Carpentier et~al.(2019{\natexlab{b}})Carpentier, Saurel, Buondonno,
  Mirabel, Lamiraux, Stasse, and Mansard]{carpentier2019pinocchio}
J.~Carpentier, G.~Saurel, G.~Buondonno, J.~Mirabel, F.~Lamiraux, O.~Stasse, and
  N.~Mansard.
\newblock The pinocchio c++ library -- a fast and flexible implementation of
  rigid body dynamics algorithms and their analytical derivatives.
\newblock In \emph{IEEE International Symposium on System Integrations (SII)},
  2019{\natexlab{b}}.

\bibitem[Pan et~al.(2012)Pan, Chitta, and Manocha]{pan2012fcl}
J.~Pan, S.~Chitta, and D.~Manocha.
\newblock Fcl: A general purpose library for collision and proximity queries.
\newblock In \emph{2012 IEEE International Conference on Robotics and
  Automation}, pages 3859--3866. IEEE, 2012.

\bibitem[Ross et~al.(2011)Ross, Gordon, and Bagnell]{Ross2011}
S.~Ross, G.~J. Gordon, and D.~Bagnell.
\newblock A reduction of imitation learning and structured prediction to
  no-regret online learning.
\newblock In G.~J. Gordon, D.~B. Dunson, and M.~Dud{\'{\i}}k, editors,
  \emph{Proceedings of the Fourteenth International Conference on Artificial
  Intelligence and Statistics, {AISTATS} 2011, Fort Lauderdale, USA, April
  11-13, 2011}, {JMLR} Proceedings, pages 627--635. JMLR.org,
\newblock 2011.

\bibitem[Ratliff et~al.(2009)Ratliff, Zucker, Bagnell, and
  Srinivasa]{Ratliff2009}
N.~D. Ratliff, M.~Zucker, J.~A. Bagnell, and S.~S. Srinivasa.
\newblock {CHOMP:} gradient optimization techniques for efficient motion
  planning.
\newblock In \emph{2009 {IEEE} International Conference on Robotics and
  Automation, {ICRA} 2009, Kobe, Japan, May 12-17, 2009}, pages 489--494.
  {IEEE},
\newblock
\newblock 2009.

\bibitem[Ichter and Pavone(2019)]{Ichter2019}
B.~Ichter and M.~Pavone.
\newblock {Robot Motion Planning in Learned Latent Spaces}.
\newblock \emph{IEEE Robotics and Automation Letters}, 2019.
\newblock
\newblock ISSN 23773766.

\end{thebibliography}

\end{document}